\documentclass{article}
\usepackage{spconf,amsmath,graphicx,amsfonts,pifont}
\usepackage{booktabs,arydshln,multirow,subcaption}
\usepackage{floatrow}
\usepackage[d]{esvect}

\DeclareMathOperator{\PE}{\vv{PE}}
\DeclareMathOperator{\PEb}{\vv{PE}_2}
\DeclareMathOperator{\PEt}{\vv{PE}_3}
\DeclareMathOperator{\px}{px}
\DeclareMathOperator{\attn}{Attention}
\DeclareMathOperator{\softm}{softmax}
\DeclareMathOperator{\conv}{Conv}
\DeclareMathOperator{\CIA}{CIA}
\DeclareMathOperator{\AM}{AM}
\newcommand{\cmark}{\ding{51}}%
\newcommand{\xmark}{\ding{55}}%

\title{DepthFormer: Multimodal Positional Encodings and Cross-Input Attention for Transformer-Based Segmentation Networks}
\name{Francesco Barbato, Giulia Rizzoli, Pietro Zanuttigh\thanks{This work has been in part supported by the University of Padova under the project "SID2020 Semantic Segmentation in the Wild". %
}}
\address{Department of Information Engineering\\
         University of Padova\\
         Via Gradenigo 6/a, Padova, Italy}
\begin{document}
\maketitle
\begin{abstract}
Most approaches for semantic segmentation use only information from color cameras to parse the scenes, yet recent advancements show that using depth data allows to further improve performances. 
In this work, we focus on transformer-based deep learning architectures, that have achieved state-of-the-art performances on the segmentation task, and we propose to employ depth information by embedding it in the positional encoding. Effectively, we extend the network to multimodal data without adding any parameters and in a natural way that exploits the strength of transformers' self-attention modules. 
We also investigate the idea of performing cross-modality operations inside the attention module, swapping the key inputs between the depth and color branches.
Our approach consistently improves performances on the Cityscapes benchmark. %
\end{abstract}
\vspace{-1em}
\section{Introduction}
\label{sec:intro}
\begin{figure*}[tbh]
    \centering
    \includegraphics[width=.75\textwidth,trim={0 1cm 0 1cm},clip]{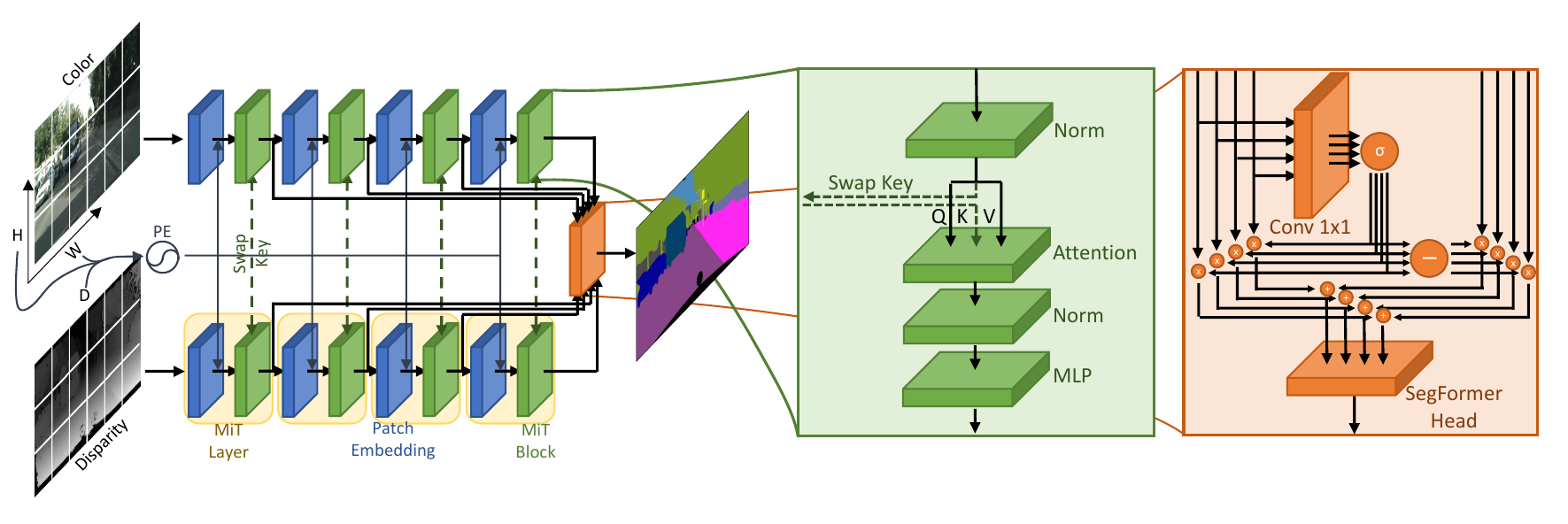}
    \caption{Graphical representation of our architecture, we highlight the Cross-Input Attention (CIA, in green) and the Attention-Mix (AM, in orange) modules. Note that the two encoder branches share weights.}
    \label{fig:architecture}
\end{figure*}

Semantic Segmentation, i.e., pixel-wise classification of an input image,  has recently become a very active %
research field, especially given the potential applications in autonomous driving. %
Among approaches for this task, architectures based on convolutional models have achieved-state-of-the art results for many years~\cite{mo2022review}. %
More recently, after being adapted to vision tasks in the ViT work~\cite{dosovitskiy2020image}, transformer-based approaches~\cite{vaswani2017attention} have started to come up. %
SegFormer~\cite{xie2021segformer}, the architecture on which our work is based on, is an example of this class. Furthermore, recent research has also shown that multimodal approaches, exploiting also depth or other side information, allow to further improve performances~\cite{technologies10040090}.

In this paper, we propose to tackle multimodal semantic segmentation with two different strategies tailored for transformer-based architectures.
Firstly, we design a way to inject depth information in the positional encoding of a transformer, effectively expanding the embedding latent dimensionality from $2$D to $3$D, with a noticeable gain in the final segmentation accuracy and with no additional parameters or significant computational cost.
Secondly, we consider the three inputs %
of a transformer self-attention head, and propose to swap such inputs between parallel architectures fed with different modality inputs. %

The article is structured as follows: in Sec.~\ref{sec:related} we report a survey of the related works; then, a detailed description of our approach is presented in Sec.~\ref{sec:method}; %
Sec.~\ref{sec:results} contains the numerical results and ablation studies %
on the Cityscapes~\cite{cordts2016cityscapes} dataset%
; finally, in Sec.~\ref{sec:conclusions} we provide  conclusive remarks and possible future perspectives.
\section{Related Works}
\label{sec:related}
After showing impressive results in text recognition, attention-based strategies %
attempting to capture long-range relationships between input data have recently been applied to vision tasks. The Vision Trasformers (ViT)~\cite{dosovitskiy2020image} %
work introduced a convolution-free vision strategy based on transformers capable of outperforming prior state-of-the-art methods in image classification. SegFormer~\cite{xie2021segformer} is a variation of the model tackling semantic segmentation. It contains a positional-encoding-free, hierarchical Transformer encoder and a lightweight all-MLP decoder.

Recent research has shown that other modalities such as depth and thermal cameras can support the extraction of semantic cues~\cite{technologies10040090,valada2020self,liu2022cmx,testolina2022selma}. The first attempts at multimodal semantic segmentation combine RGB data with other modalities into multi-channel representations, which are subsequently fed into traditional semantic segmentation networks built on the encoder-decoder framework~\cite{couprie2013indoor,pagnutti17}. This straightforward early fusion strategy struggles to capture the different types of information carried by the various modalities %
and, as a result, it is not very effective. The majority of current methods attempt to combine the early, feature, and late-fusion strategies in the best possible way by performing several fusion operations at various levels in the deep network~\cite{valada2020self,zhang2021abmdrnet,krispel2020fuseseg}.
A popular architectural choice is using a multi-stream encoder with a network branch processing each modality %
together with additional network modules connecting the different branches that combine modality-specific features into fused ones and/or carry information across the branches~\cite{valada2020self,liu2022cmx,chen2020bi}. This hierarchical fusion technique creates a more precise feature map by utilizing multilevel features through incremental feature merging. Instead of only at early or late stages, this approach requires fusing information at multiple levels.
Multi-level features can be supplied in this way either mutually between modalities \cite{valada2020self,liu2022cmx} or in one-way, as in~\cite{seichter2021efficient}, where depth cues are sent to the RGB branch.
\section{Proposed Method}
\label{sec:method}

In this section, we  present a detailed description of our approach (see Fig.~\ref{fig:architecture}), starting from introducing the employed framework and then %
detailing the proposed $3$D Positional Encoding (PE) and cross-attention strategies. 
Our approach %
moves from the recent and well-performing SegFormer~\cite{xie2021segformer} architecture, whose backbone has been modified to make use of Positional Encodings %
instead of the Mix-FFN layer. The PE is injected before each processing block, similarly to the Pyramid Vision Transformer~\cite{wang2021pyramid} backbone.
To account for the multimodal input, we adapted the SegFormer architecture in a dual-stream version, where the depth information (represented through disparity data) is forwarded to the Positional Encoding and to one of the two branches as well. %
Compared to previous research we introduce two key achievements, a novel positional encoding scheme (Sec.~\ref{subsec:positional}) and a multi-modal attention scheme (Sec.~\ref{subsec:cross}).
\subsection{3D Positional Encoding}
\label{subsec:positional}
\begin{figure}[tbh]
    \centering
    \includegraphics[width=.85\textwidth]{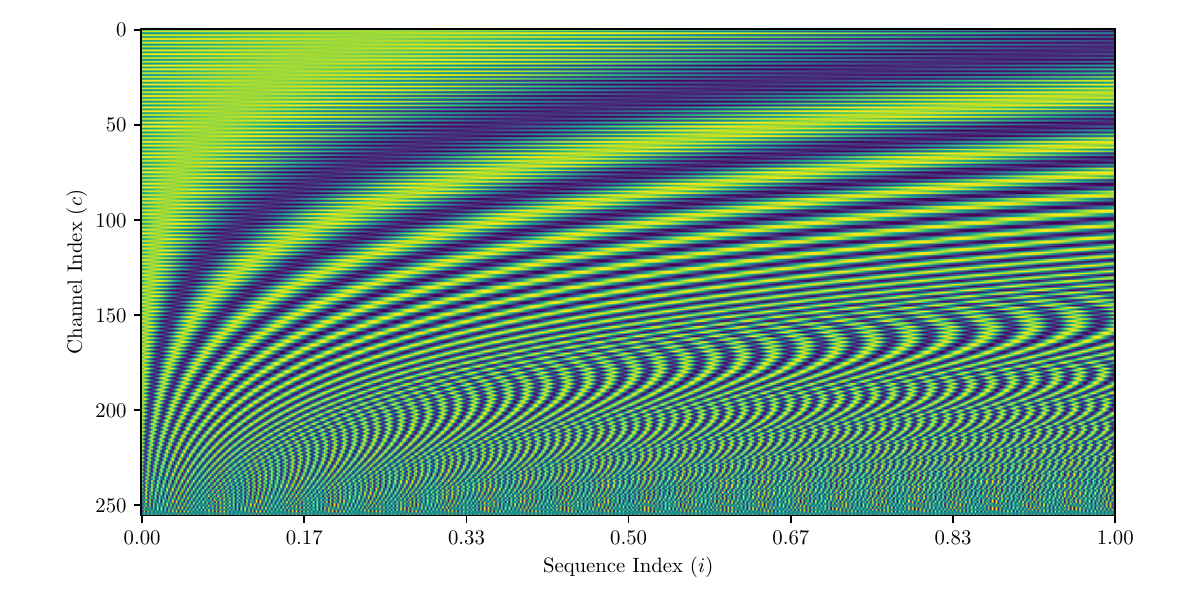}
    \caption{Graphical representation of the $\PE$ embeddings.}
    \label{fig:PE_1d}
\end{figure}

The starting point for the design of our positional encoding is the scheme presented in \cite{vaswani2017attention}, which encodes the position of a token in a sequence using a multidimensional function $\PE:~\mathbb{R}\mapsto\mathbb{R}^C$. We chose this type of encoding, rather than a learned one because it accepts arbitrary 1-dimensional inputs and can be used to embed the depth information. 
More in detail, we used the following encoding function:
\begin{equation}
    \PE(i)[c] = \begin{cases}
                    \sin(i \pi2^{\log_2{I}\;c/C}) & \text{if  } c  \text{ mod } 2 = 0   \\
                    \cos(i \pi2^{\log_2{I}\;c/C}) & \text{if  } c  \text{ mod }  2  = 1  
                \end{cases}
    \label{eq:pe1d}
\end{equation}
where the position $i$ is normalized in the range $[0,1]$, $I$ is the maximum value of $i$ before normalization and $C$ is the number of dimensions (note that $c$ is 0-indexed).
Notice that in \cite{vaswani2017attention} the frequencies used in the embedding reach up to $10^5$, to account for long input sequences in natural language processing. In our case, the sequence length is the image side (i.e., $512$px) therefore we change the scaling factor accordingly. %
See Fig.~\ref{fig:PE_1d} for a graphical representation of the  encoding.

\begin{figure}[tbh]
    \centering
    \includegraphics[width=.75\textwidth]{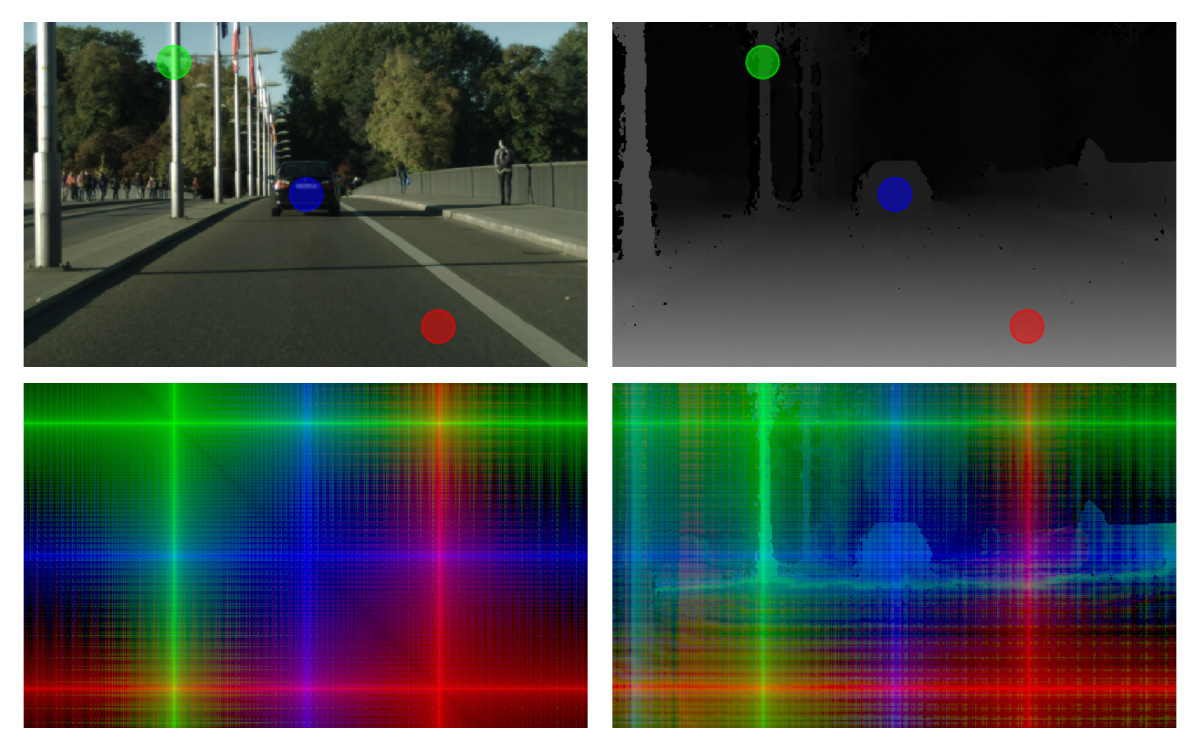}
    \caption{Comparison between $2$D (left) and $3$D (right) positional encoding. The pixels highlighted in the two top images are used as targets: the distance between them and all other pixels is represented in the bottom images.}
    \label{fig:PE_compare}
\end{figure}

To handle multidimensional sequences (images) we needed to extend the definition of $\PE$ to multiple dimensions and heterogeneous inputs (pixel coordinates $u$ and $v$ and depth/disparity $d$). We settled on summing together the embedding of each dimension: %
\begin{align}
    \label{eq:pe2d}
    \PEb(u_n,v_n) &= \PE(u_n)+\PE(v_n) \\
    \label{eq:pe3d}
    \PEt(u_n,v_n,d_n) &= \PE(u_n)+\PE(v_n)+\PE(d_n)
\end{align}
This implies that all directional encodings contribute equally to the final token representation, avoiding directional biases.
A visual representation of the difference between $\PEb$ and $\PEt$ is reported in Fig.~\ref{fig:PE_compare}, where three target pixels (red, green, blue) are sampled from an input image (top row) and the similarity in embedding between them and each other pixel is reported in the bottom two images using the same colors. The image in the bottom left shows similarity scores according to $\PEb$, while the bottom right the ones according to $\PEt$. One can easily appreciate the much richer description of the content offered by the three-dimensional encoding, which captures the stronger relation between pixels belonging to the same object. Notice, for example, how the car in the center (blue target pixel) is clearly identifiable in the similarity representation. 
\vspace*{-1em}
\subsection{Multimodal Attention}
\label{subsec:cross}
For the multimodal setup, we exploit a shared-weights parallel branches encoder.
The relevant information is shared between the two modalities using two modules: a Cross-Input Attention (CIA) block and an Attention-Mix (AM) one.
To better describe the CIA block, we report the original formulation of a Transformer self-attention module and show how it can be modified:
\begin{equation}
    \label{eq:attn}
    \attn(Q,K,V) = \softm(\frac{QK^{T}}{\sqrt{d_{head}}})V
\end{equation}
where $Q$ is the query, $K$ is the key, $V$ is the value and $d_{head}$ is the dimension of the head \cite{vaswani2017attention}.
The attention operation can be thought of as a retrieval process, where a generic query $Q$, specified by a key $K$ is used to retrieve a value $V$. 

This concept leads to the idea of swapping the self-attention tuples ($Q$, $K$, $V$) between the color and depth branches in order to allow for retrievals spanning different sources. %
Given that the key is what specifies the values to be retrieved, it is reasonable to assume that swapping it between modalities provides mutual complementary information to each attention head.
This was confirmed by experimental studies, where we observed that swapping the key values led to the best performances.  %
The %
Cross-Input Attention module can be expressed in mathematical notation as:
\begin{equation}
    \label{eq:cross_attn}
    \begin{array}{l}
        \CIA_c = \attn(Q_{c},K_{d},V_{c}) \\
        \CIA_d = \attn(Q_{d},K_{c},V_{d})
    \end{array}
\end{equation}
where $c$ and $d$ represent color and depth (disparity), respectively.
In such a manner, the values are re-weighted by the relevance between the queries and the keys of the two different modalities, enhancing features exchange.

Up to this point, the two modalities were processed independently, but they need to be merged before being fed to the SegFormer~\cite{xie2021segformer} classifier. 
Previous research showed that re-aligning multi-resolution outputs between the two modalities would improve consistency \cite{zhang2022central}, therefore we design the Attention-Mix (AM) module with this goal.
The module processes each MiT~\cite{xie2021segformer} layer output $O_{[c,d],[1,2,3,4]}$ before it is fed to the segmentation head in the following manner:
\begin{equation}
    \label{eq:attn_mix}
    \begin{split}
        \AM(O_{c,\cdot}, O_{d,\cdot}) \; = \; &O_{c,\cdot} \sigma(\conv(O_{c,\cdot})) \;\; + \\ &O_{d,\cdot}(1-\sigma(\conv(O_{c,\cdot})))
    \end{split}
\end{equation}
Where $\conv$ is a $1\times1$ convolution, $\sigma$ represents the sigmoid function and the multiplication is performed point-wise.
See Fig.~\ref{fig:architecture} %
(rightmost part) for a visual scheme of this module.

\section{Results}
\label{sec:results}

\begin{table*}[htbp]
  \RawFloats
\begin{minipage}{0.73\textwidth}
\caption{Per class IoU results on the Cityscapes~\cite{cordts2016cityscapes} dataset, best in \textbf{bold}. (r) means that the original code was re-run using our backbone; $\dagger$ means that the fusion was performed with an ad-hoc strategy, + fusion by feature sum, * attn-mix fusion.}
\label{tab:results}
\linespread{0.7}\selectfont\centering
\resizebox{1\textwidth}{!}{
\renewcommand{\tabcolsep}{0.08cm}
\begin{tabular}{cccccccccccccccccccccccc}
\toprule[1pt]
& \hspace*{.75em}Configuration & \rotatebox{90}{\parbox{30pt}{\centering Dual-Branch}} & \rotatebox{90}{Fusion}\hspace*{.75em} & \rotatebox{45}{Road}\hspace*{-1em} & \rotatebox{45}{Sidewalk}\hspace*{-2em} & \rotatebox{45}{Building}\hspace*{-2em} & \rotatebox{45}{Wall}\hspace*{-1em} & \rotatebox{45}{Fence}\hspace*{-1em} & \rotatebox{45}{Pole}\hspace*{-1em} & \rotatebox{45}{Traffic Light}\hspace*{-3em} & \rotatebox{45}{Traffic Sign}\hspace*{-3em} & \rotatebox{45}{Vegetation}\hspace*{-2em} & \rotatebox{45}{Terrain}\hspace*{-1em} & \rotatebox{45}{Sky}\hspace*{-.5em} & \rotatebox{45}{Person}\hspace*{-1em} & \rotatebox{45}{Rider} & \rotatebox{45}{Car} & \rotatebox{45}{Truck}\hspace*{-1em} & \rotatebox{45}{Bus}\hspace*{-.5em} & \rotatebox{45}{Train}\hspace*{-1em} & \rotatebox{45}{Motorbike}\hspace*{-2em} & \rotatebox{45}{Bicycle} & \hspace*{.75em}\makebox{mIoU}\hspace*{.75em} \\
\noalign{\smallskip} 
\toprule[1pt]
& RGB Baseline & \xmark & - & 97.5 & 83.2 & 90.7 & 47.8 & 56.0 & 44.8 & 56.5 & 67.2 & 91.4 & 68.4 & 94.5 & 78.9 & 49.6 & 93.3 & 68.0 & 71.1 & 73.7 & 57.8 & 70.1 & 71.6 \\
& Depth Baseline & \xmark & - & 95.7 & 72.8 & 84.0 & 27.9 & 36.4 & 31.4 & 34.0 & 39.1 & 80.0 & 49.2 & 83.3 & 60.4 & 27.3 & 85.1 & 45.9 & 49.8 & 37.5 & 32.9 & 47.7 & 53.7 \\
& CMX~\cite{liu2022cmx} (r) & \cmark & $\dagger$ & 97.2 & 78.6 & 90.2 & \textbf{53.2} & 50.0 & 52.4 & 56.8 & 63.7 & 91.0 & 62.1 & 94.5 & 72.3 & 44.6 & 90.9 & 50.9 & 65.5 & 19.0 & 37.9 & 64.6 & 65.0 \\
\hdashline
\multirow{6}{*}{\rotatebox{90}{Ours}} & RGBD & \cmark & +  & 97.5 & 83.3 & 91.2 & 45.9 & 55.6 & 47.8 & 59.8 & 69.2 & 91.4 & 68.4 & 94.6 & 79.4 & 48.5 & 93.4 & 69.2 & 71.1 & 74.8 & 57.6 & 70.1 & 72.0 \\
& $3$D PE & \xmark & - & 97.5 & 83.3 & 90.9 & 46.2 & 55.7 & 46.1 & 57.5 & 67.7 & 91.4 & 68.5 & 94.7 & 78.6 & 49.7 & 93.3 & 71.3 & 72.8 & \textbf{76.7} & 57.7 & 69.7 & 72.1 \\
& cross-$V$ & \cmark & + & 97.5 & 83.4 & 91.1 & 44.8 & 55.7 & 47.8 & 59.7 & 69.3 & 91.5 & 68.4 & 94.6 & 79.5 & 48.7 & 93.4 & 69.0 & 71.1 & 73.7 & 57.7 & 70.0 & 71.9\\
& cross-$Q$ & \cmark & + & 97.5 & 83.3 & 91.2 & 45.2 & 57.2 & 48.1 & 60.0 & 69.2 & 91.5 & 67.8 & 94.6 & 79.5 & 49.4 & 93.4 & 69.2 & 71.3 & 75.3 & 57.4 & 70.4 & 72.2 \\
& cross-$K$ & \cmark & + & 97.5 & 83.4 & 91.2 & 45.9 & 56.7 & 47.7 & 59.7 & 69.3 & 91.4 & 68.4 & 94.6 & 79.6 & 49.2 & 93.4 & 69.9 & 71.9 & 75.3 & 58.1 & 70.2 & 72.3\\
& attn-mix & \cmark & * & 97.5 & 83.2 & 91.3 & 45.9 & 56.7 & 48.4 & 60.1 & 69.3 & 91.4 & 68.2 & 94.6 & 79.9 & 50.7 & 93.5 & 70.8 & 72.2 & 77.0 & 57.1 & 70.8 & 72.6 \\
& Total & \cmark & * & \textbf{97.7} & \textbf{84.3} & \textbf{91.8} & 43.2 & \textbf{58.7} & \textbf{52.0} & \textbf{63.1} & \textbf{72.0} & \textbf{91.9} & \textbf{69.6} & \textbf{94.8} & \textbf{81.0} & \textbf{52.8} & \textbf{94.0} & \textbf{71.5} & \textbf{74.4} & \textbf{76.7} & \textbf{60.4} & \textbf{71.8} & \textbf{73.8} \\
\bottomrule
\end{tabular}
}
\end{minipage}
\hspace{8px}
\begin{minipage}{0.25\textwidth}
    \centering
    \caption{Comparison between our approach and competitors with backbones of similar complexity.}
    \label{tab:compare}
    \resizebox{1\textwidth}{!}{
    \begin{tabular}{ccc}
        \toprule
        Method & Backbone & mIoU \\
        \midrule
        CMFNet~\cite{zhang2022central} & VGG16 & 60.0 \\
        MDASS~\cite{rashed2019motion} & VGG16 & 63.1 \\
        CMX~\cite{liu2022cmx} (r) & MiT-B0 (PE) & 65.0 \\
        RSSAWC~\cite{https://doi.org/10.48550/arxiv.1905.10117} & ICNet & 65.1 \\
        RTFNet~\cite{sun2019rtfnet} & ResNet18 & 69.4 \\
        DABNet~\cite{li2019dabnet} & DABNet & 70.1 \\
        TWTISS~\cite{hoyer2021three} & ResNet101 & 71.2 \\
        AdapNet~\cite{valada2017adapnet} & AdapNet & 71.7 \\
        DepthFormer & MiT-B0 (PE) & \textbf{73.8}\\
        \bottomrule
    \end{tabular}
    }
\end{minipage}
\end{table*}
\newcommand{\imSize}{.16\textwidth}
\begin{figure*}[h!]
\begin{subfigure}{\textwidth}
\centering
\begin{subfigure}{\imSize}
\includegraphics[width=\textwidth]{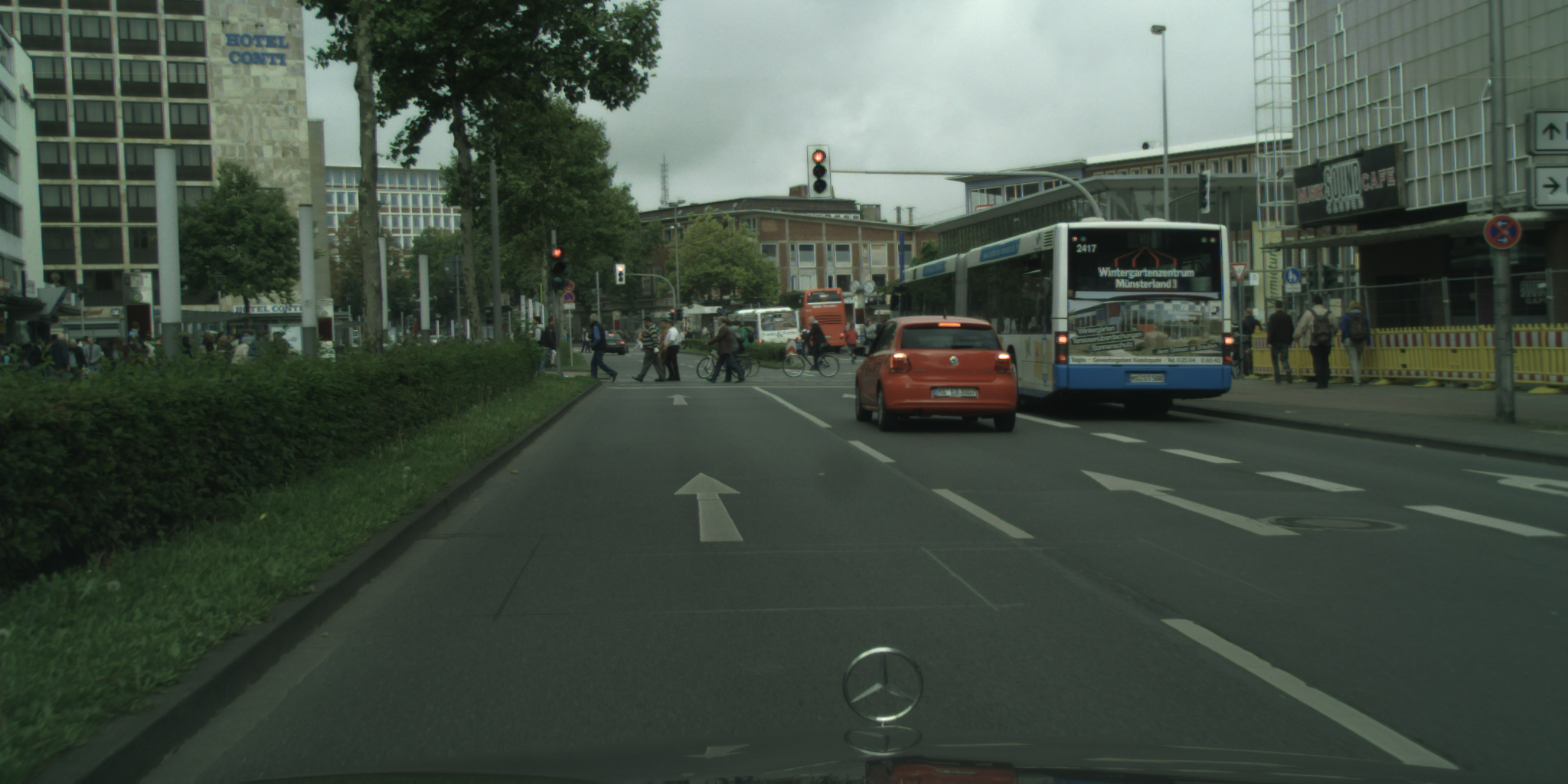}
\end{subfigure} 
\begin{subfigure}{\imSize}
\includegraphics[width=\textwidth]{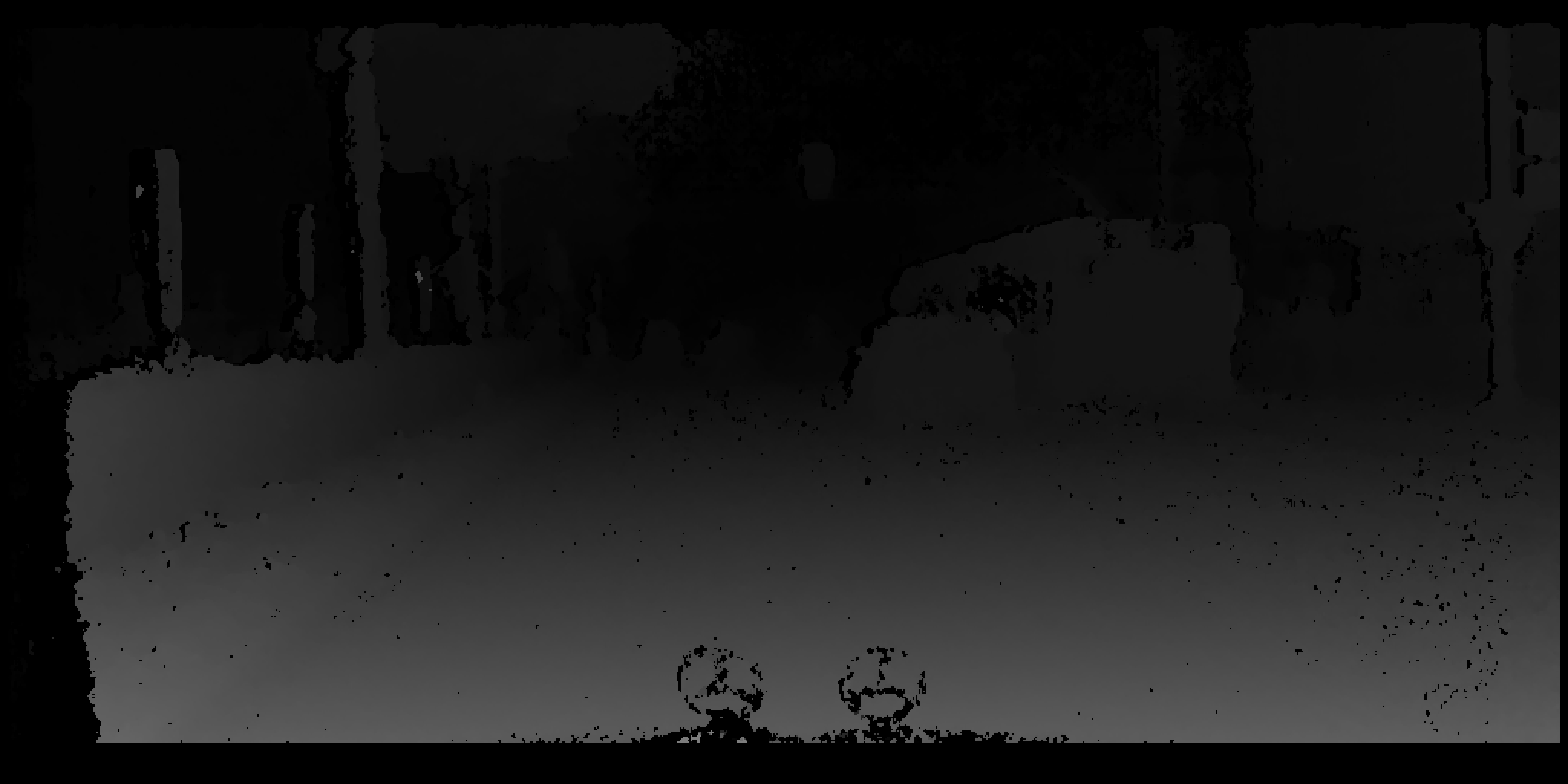}
\end{subfigure} 
\begin{subfigure}{\imSize}
\includegraphics[width=\textwidth]{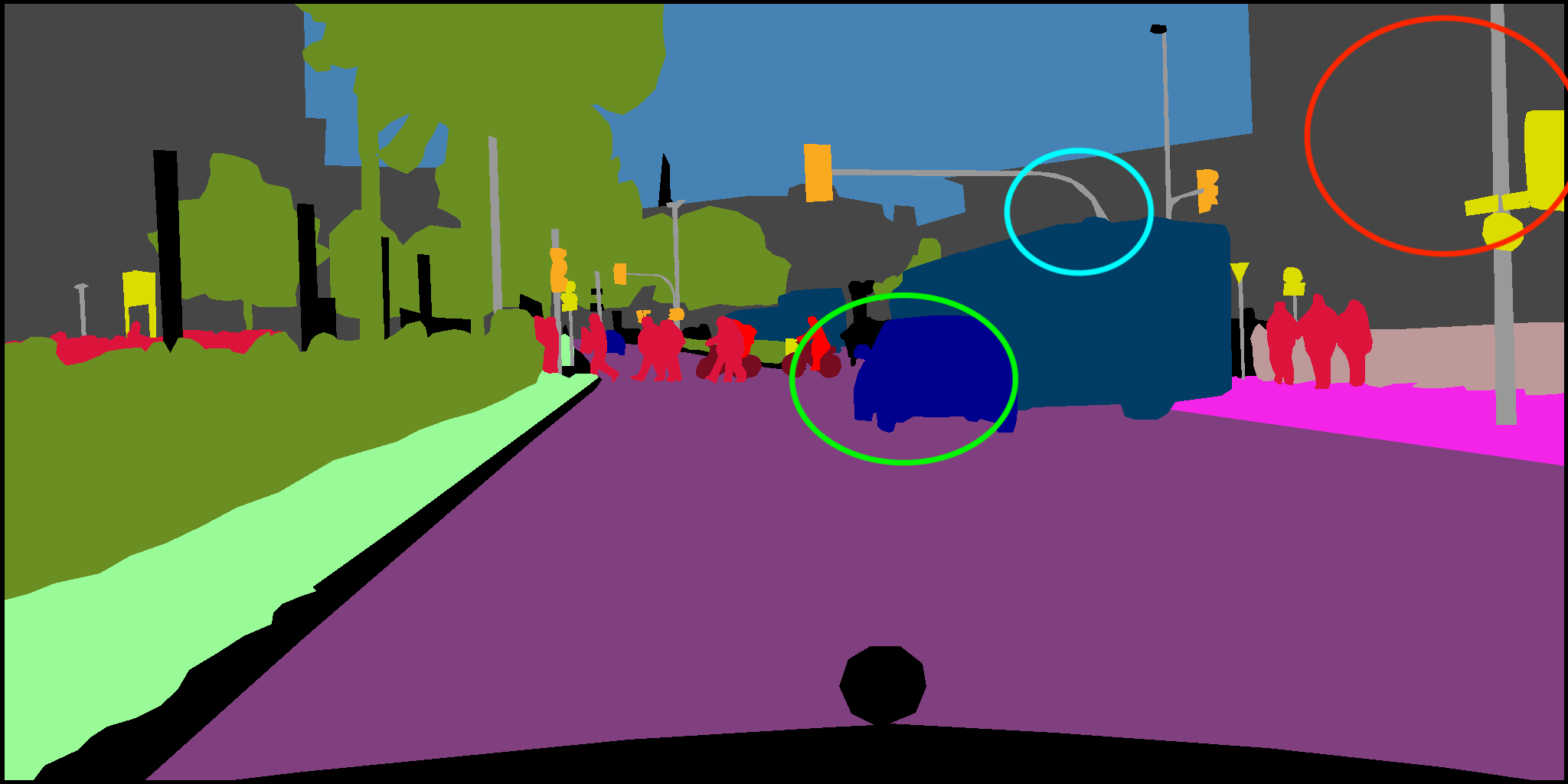}
\end{subfigure}
\begin{subfigure}{\imSize}
\includegraphics[width=\textwidth]{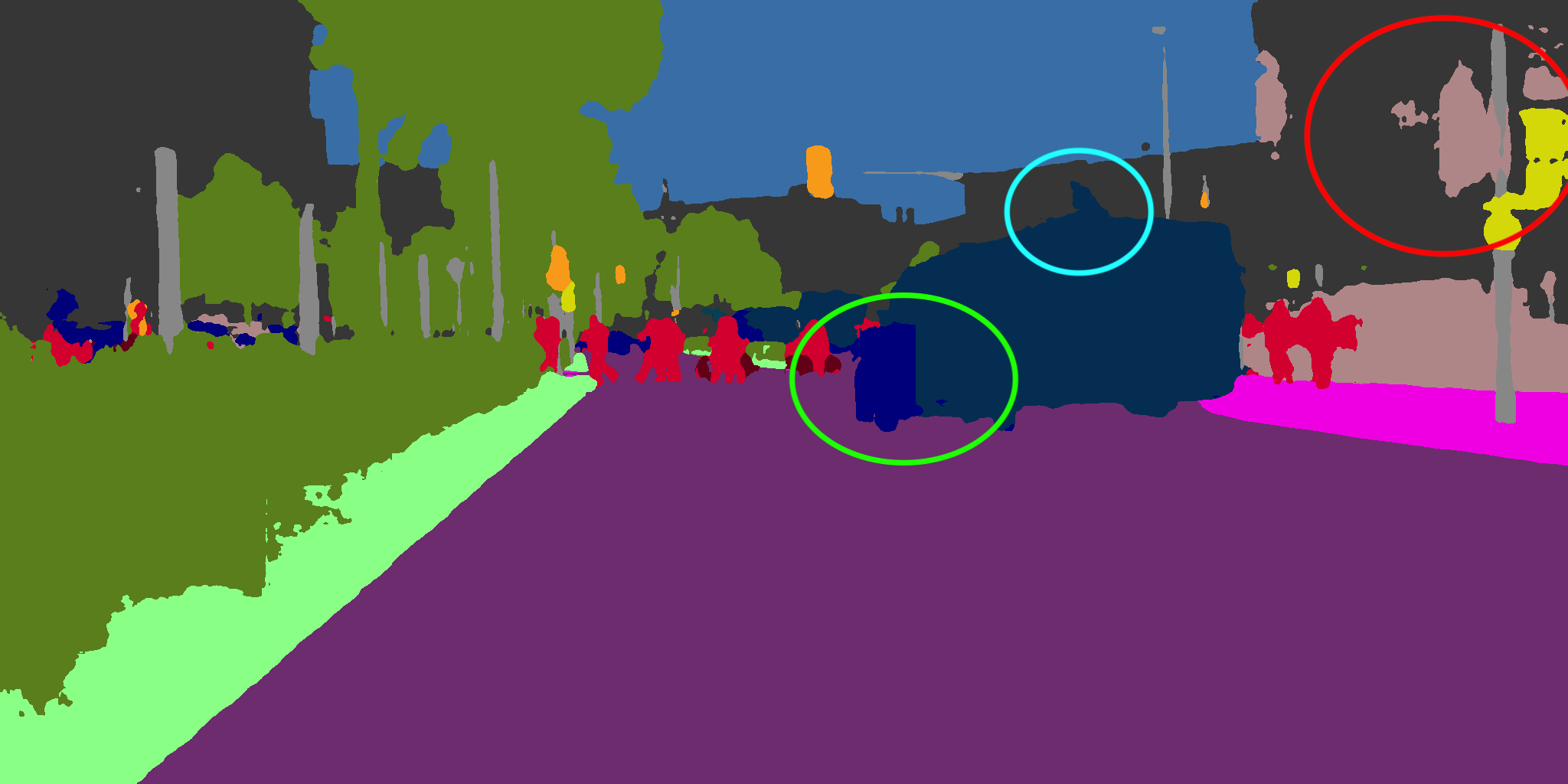}
\end{subfigure}
\begin{subfigure}{\imSize}
\includegraphics[width=\textwidth]{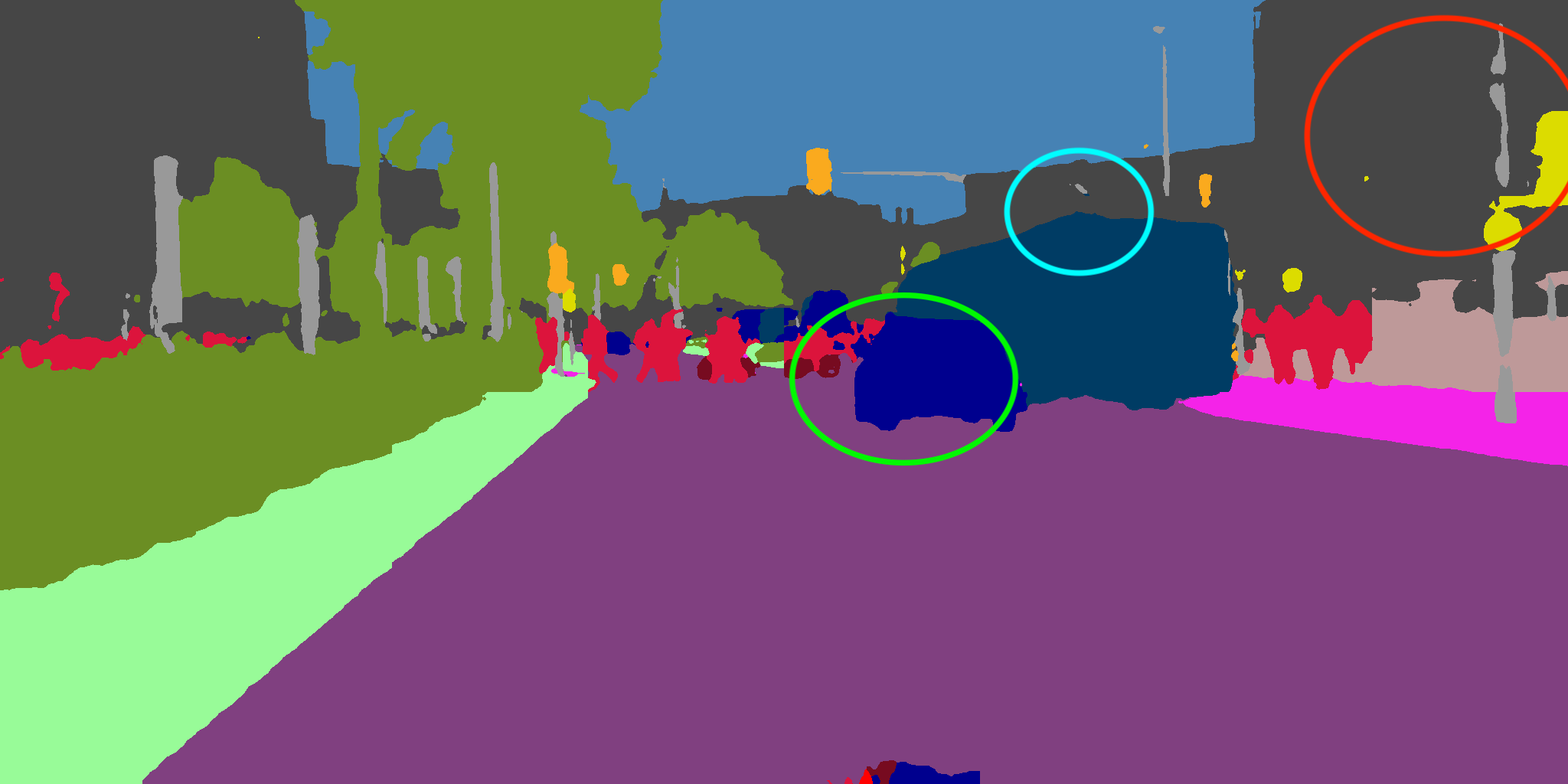}
\end{subfigure}
\begin{subfigure}{\imSize}
\includegraphics[width=\textwidth]{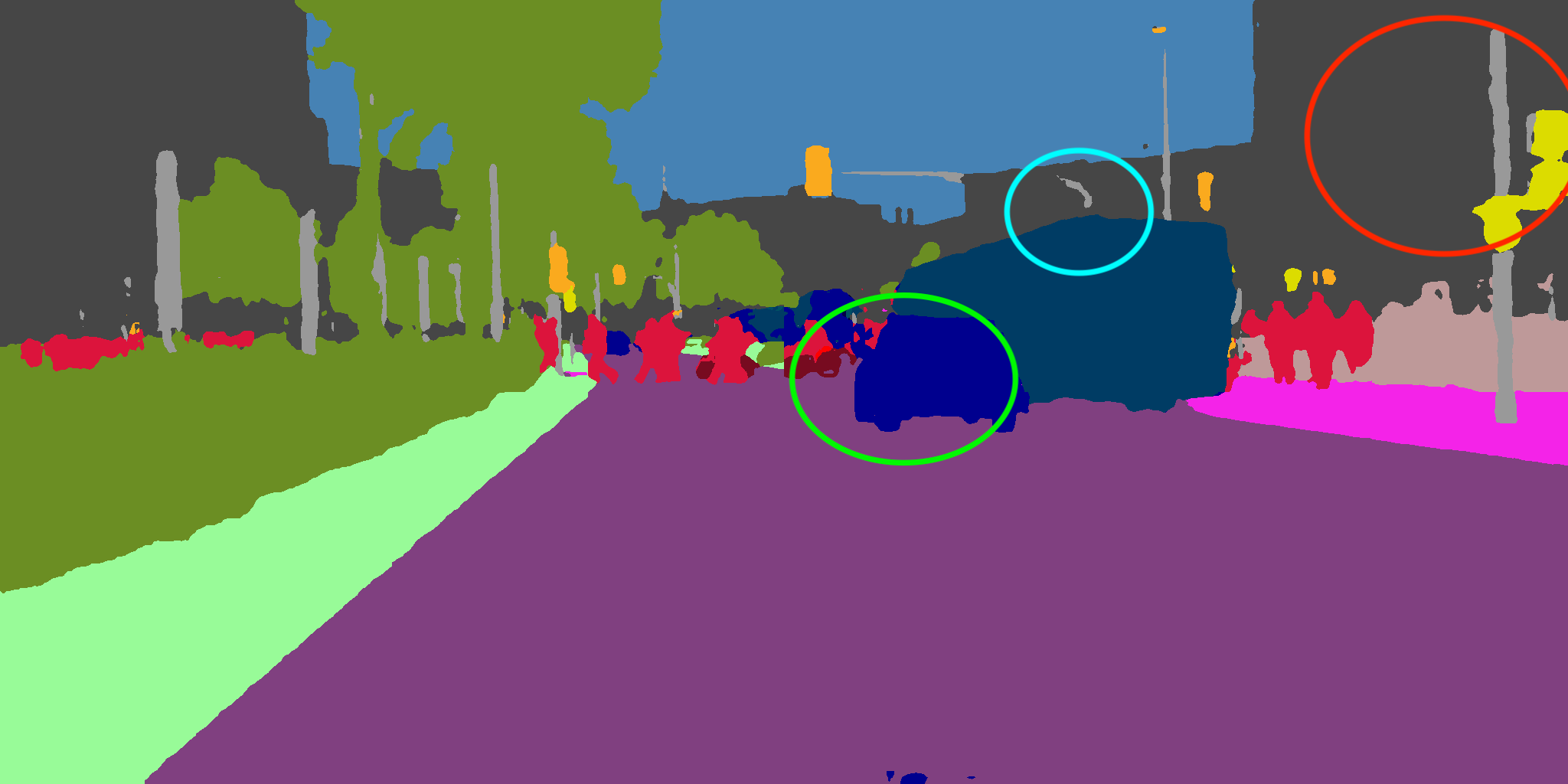}
\end{subfigure}
\end{subfigure}
\begin{subfigure}{\textwidth}
\centering
\begin{subfigure}{\imSize}
\includegraphics[width=\textwidth]{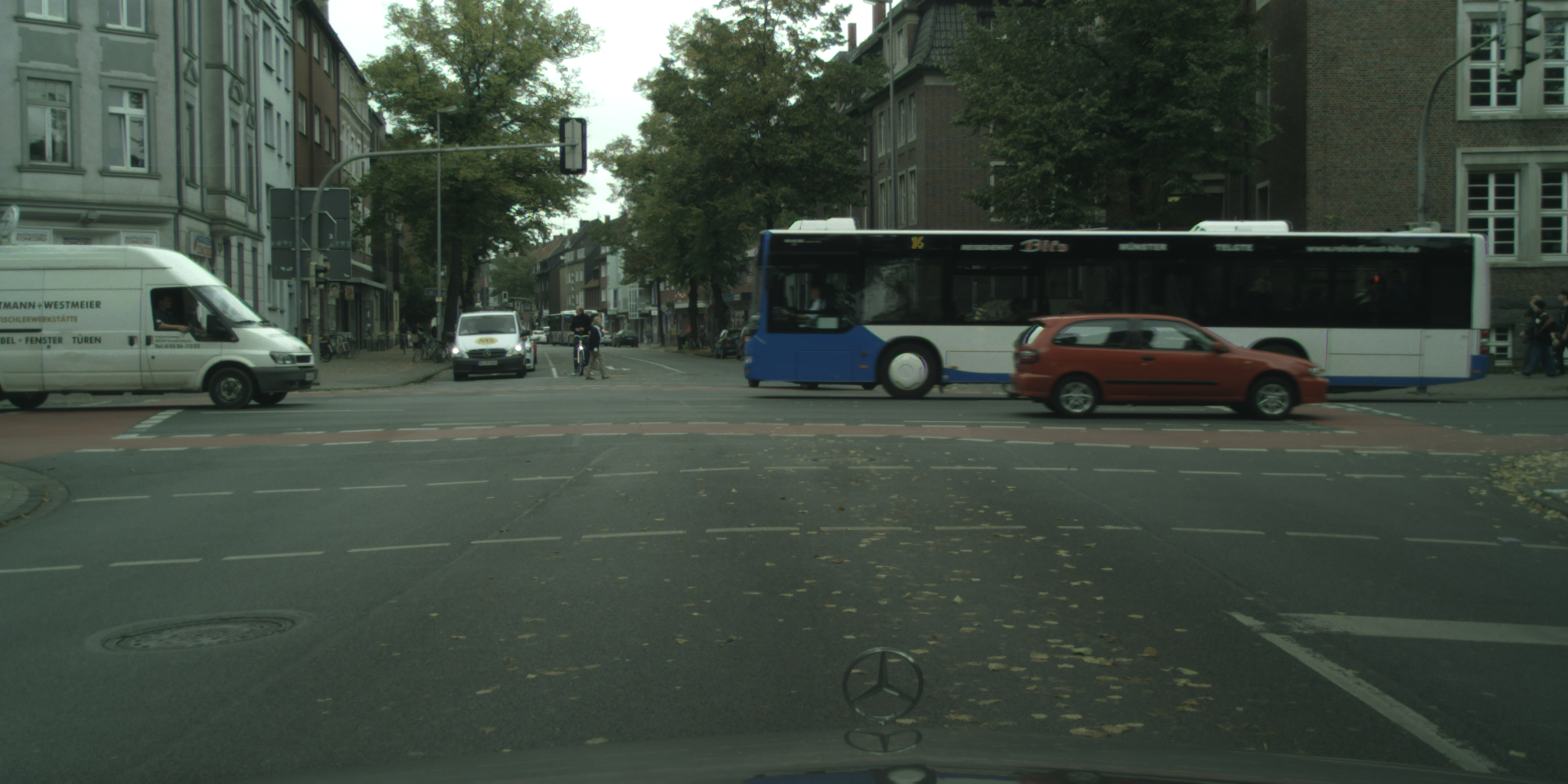}
\caption*{RGB}
\end{subfigure} 
\begin{subfigure}{\imSize}
\includegraphics[width=\textwidth]{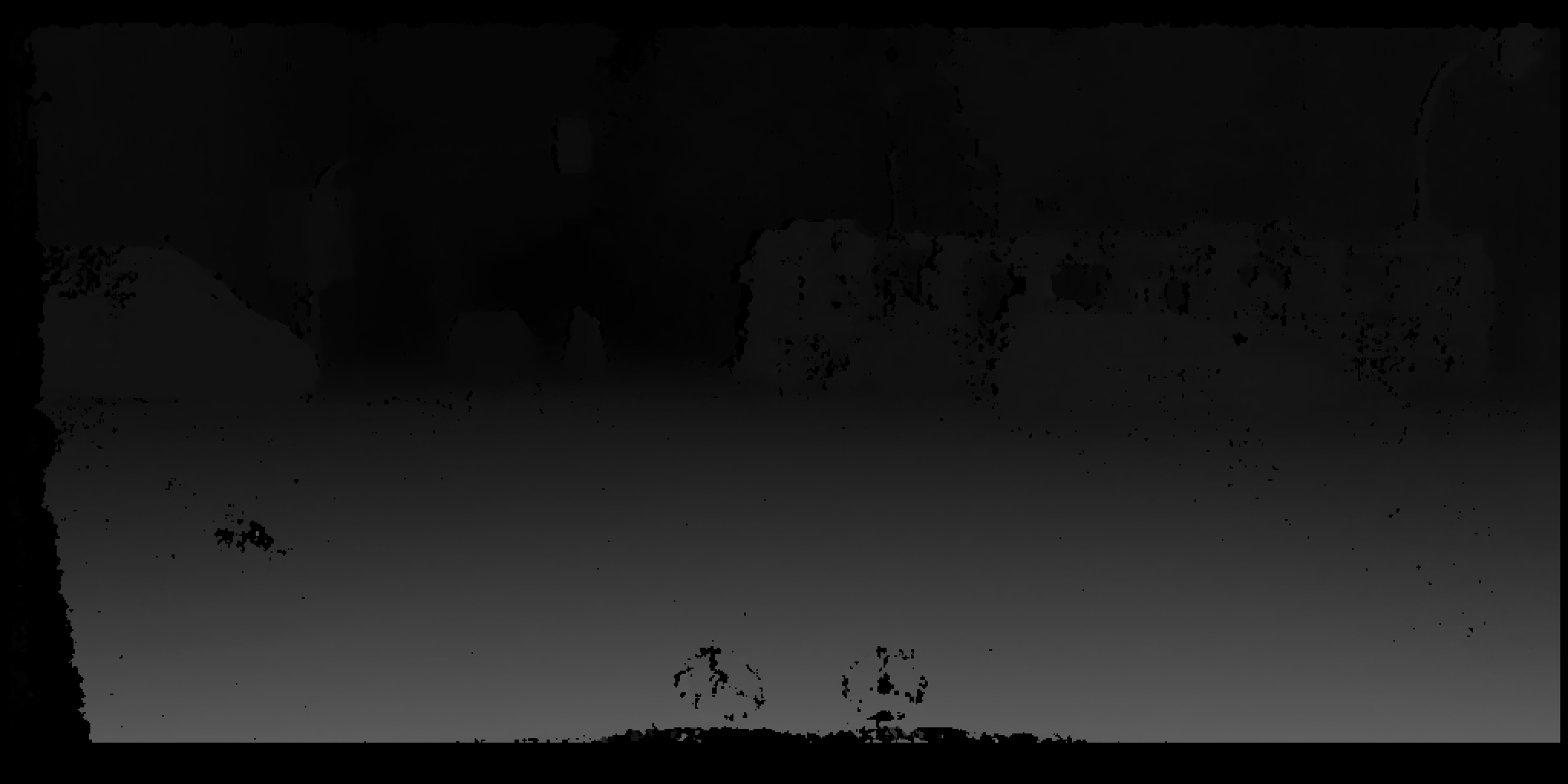}
\caption*{Depth}
\end{subfigure} 
\begin{subfigure}{\imSize}
\includegraphics[width=\textwidth]{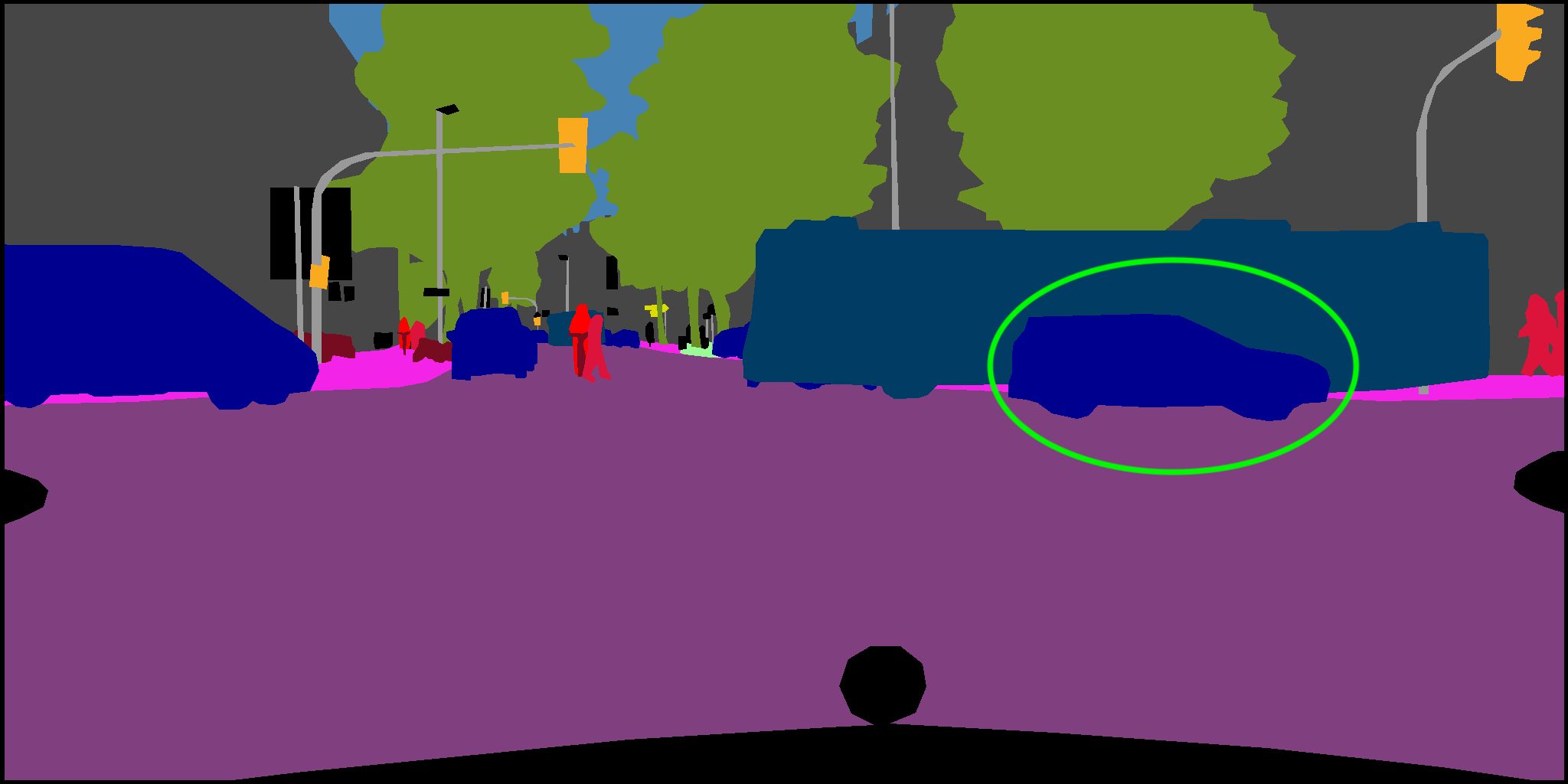}
\caption*{GT}
\end{subfigure}
\begin{subfigure}{\imSize}
\includegraphics[width=\textwidth]{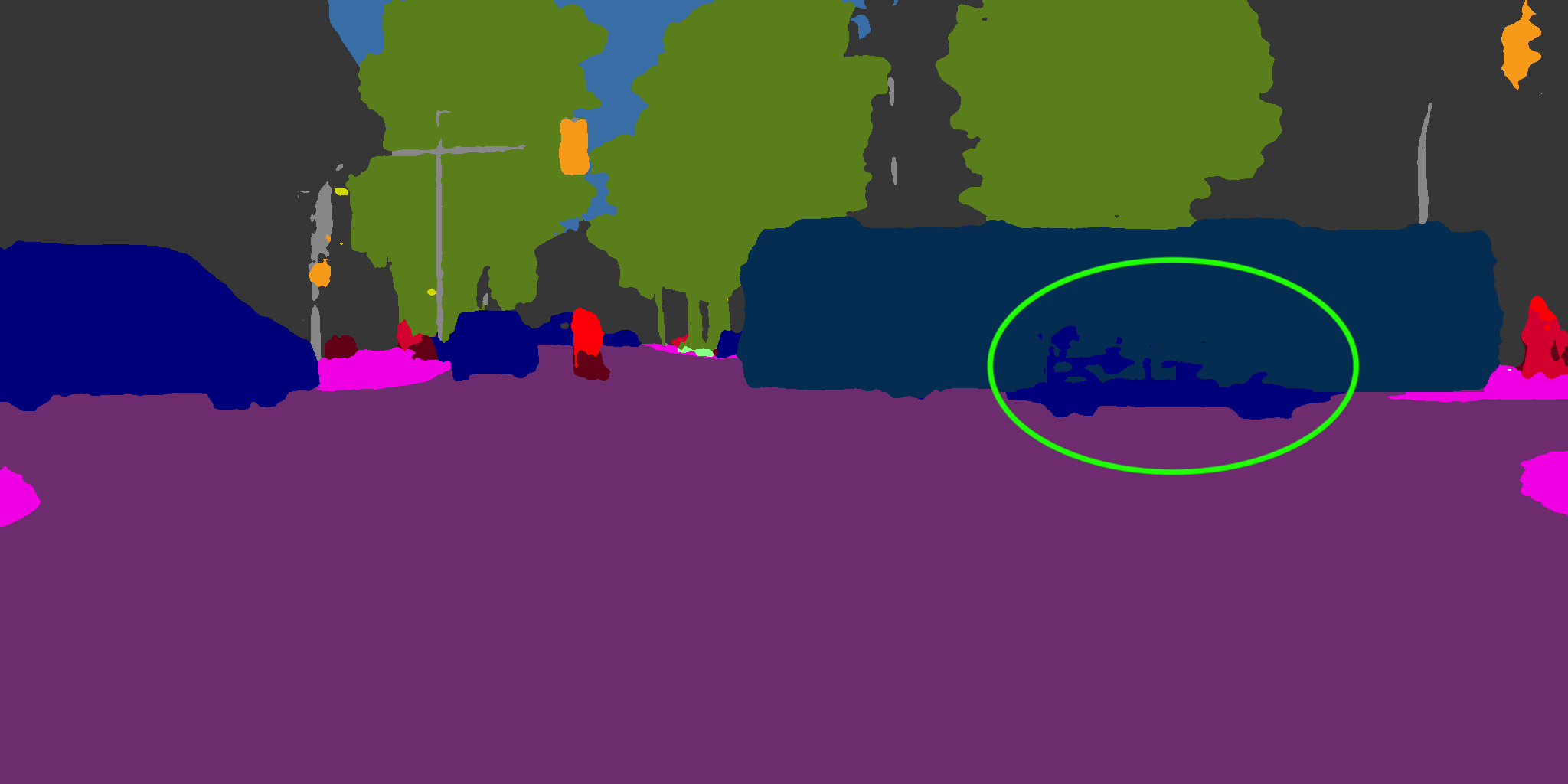}
\caption*{CMX}
\end{subfigure}
\begin{subfigure}{\imSize}
\includegraphics[width=\textwidth]{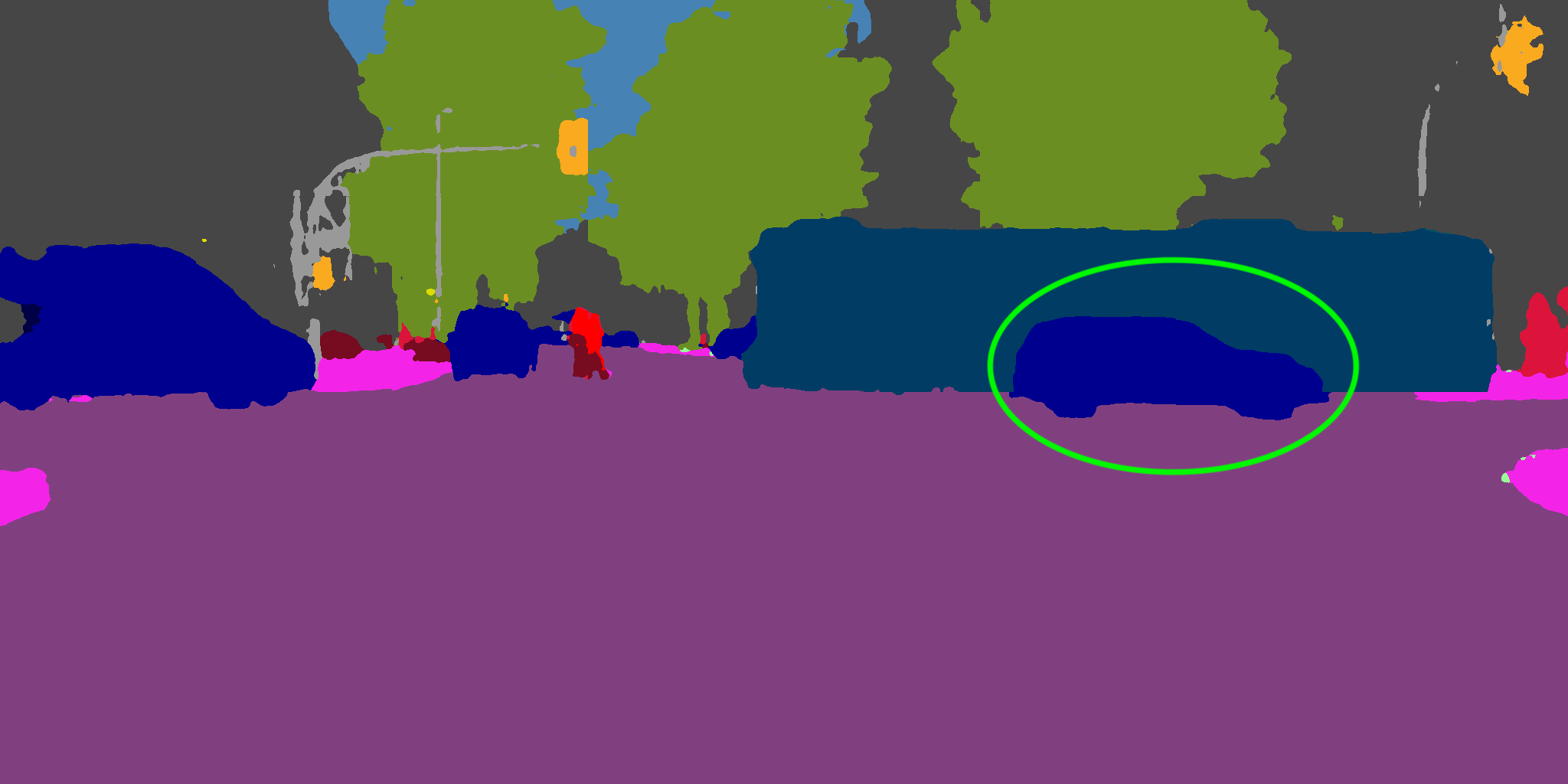}
\caption*{Ours (3D PE)}
\end{subfigure}
\begin{subfigure}{\imSize}
\includegraphics[width=\textwidth]{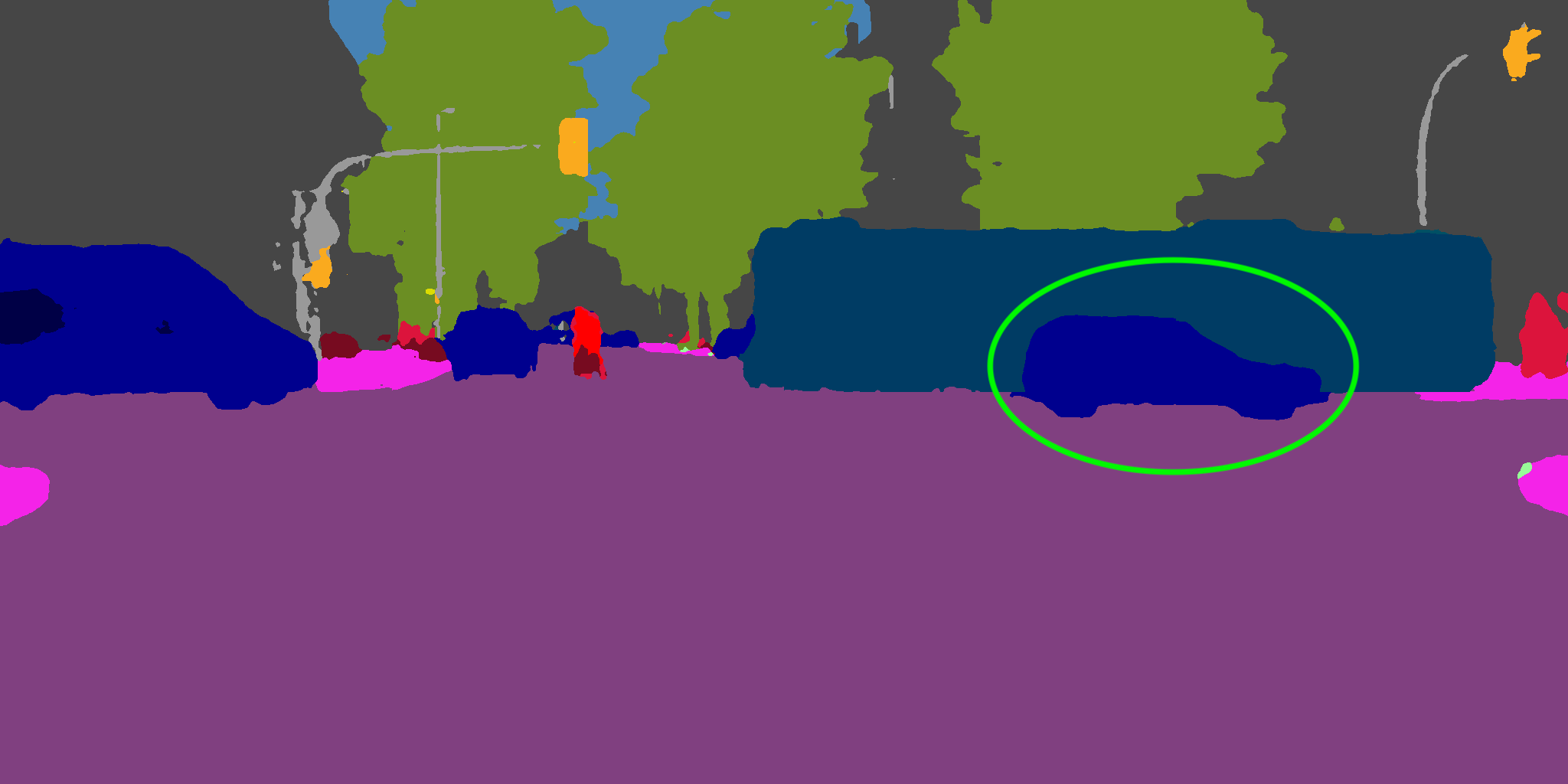}
\caption*{Ours (Total)}
\end{subfigure}
\end{subfigure}
\caption{Qualitative results of our method, for comparison we report also CMX~\cite{liu2022cmx} and the \textit{3D PE} approach.}
\label{fig:quali}
\end{figure*}

In this section, we present the numerical results attained by our approach on the Cityscapes~\cite{cordts2016cityscapes} dataset. 
We trained our architecture on Cityscapes using the Adam optimizer for $320k$ iterations on a single RTX 3060 GPU with batch 8, starting from weights pre-trained on ImageNet. The input resolution is $512\times512 \px$, and the inference is performed using sliding windows.  We follow the learning rate and data augmentation strategies of \cite{xie2021segformer} and set the weight decay rate to $0.01$.
Per-class IoU results are reported in Table~\ref{tab:results} for a color-only baseline, a depth-only baseline, for different configurations of our strategy, and for the transformer-based CMX~\cite{liu2022cmx} architecture. 
From the two baselines, it is possible to see how the exploitation of depth data is more challenging than color information, especially due to the introduction of noise and artefacts during the depth estimation, which prevents the network from fitting correctly to the data.
Nevertheless, a standard two-branches architecture (RGBD) allows improving performance  (up to $72.0\%$), which indicates the potential of using depth data and suggests that the information provided by the two modalities is complementary.

Notice how the proposed positional encoding achieves better performance than the two branches scheme, at the same time avoiding the extra complexity of the second branch (i.e., with the same complexity as the color-only architecture). 
Furthermore, CMX's two-stream architecture is not able to reach good performance when a positional encoder is employed. 

When one considers the parallel shared-weights configuration (see Fig.~\ref{fig:architecture}) it is possible to see how both our contributions %
improve performances 
(see the ablation for more details). Still, the best results are achieved by the combination of all the proposed strategies, which allows reaching an accuracy of $73.8\%$, with a noticeable gain over the color baseline even in the presence of noisy %
depth data, which makes the segmentation task more challenging. Notice  that the improvement is consistent across all classes: our approach has the highest per-class accuracy on all classes except the \textit{Wall} one. %

Furthermore, in Table \ref{tab:compare} we report some additional comparisons between our approach and competing works with a similar computational cost.
It is possible to see how our approach outperforms both convolutional architectures and transformer-based ones of comparable complexity. In particular, VGG16-based schemes achieve the worst results with a gap with respect to our approach of more than $10.0\%$ mIoU. Other multimodal strategies with lightweight backbones (ResNet18, Adapnet) allow for better performances but not enough to reach our approach. %
Importantly, CMX \cite{liu2022cmx} offers a more direct comparison given that it adapts the same architecture of our approach, i.e., SegFormer \cite{xie2021segformer}. It has been tested using the same backbone on the same RGBD data and the outcomes we obtained show that our improvements allow for much better performances.
Finally, in Fig.~\ref{fig:quali} we report some qualitative results comparing our approach  with CMX \cite{liu2022cmx} and with the \textit{3D PE} version, the circles indicate regions of interest for the comparison. Notice how CMX tends to misclassify vehicles close one to the other, such behavior is noticeable in both samples reported where the \textit{cars} in front of the \textit{trams} get confused with the latter.
\vspace{-1em}
\subsection{Ablation}
\label{subsec:ablation}
\vspace{-.5em}
Finally, we analyze the impact of all components of %
our approach. Numerical results are reported in Table \ref{tab:results}, note that the ablation experiments were trained for $160k$ iterations for speed/efficiency tradeoffs.
With regards to the positional encoding (Sec.~\ref{subsec:positional}), %
our approach shows an improvement of $0.5\%$ of mIoU w.r.t. the color baseline, without any increase in the number of architecture parameters or significant computational costs. %
The cross-input attention (Sec.~\ref{subsec:cross}) %
is analyzed through the cross-$Q$/$K$/$V$ simulations. %
Here we can see how the swap of the values leads to small but noticeable performance degradation over the RGBD baseline. %
This corresponds to a naive approach which swaps attention weights between modalities since in our shared-weight setting swapping both $K$ and $Q$ (i.e., the attention weights) is equivalent to swapping $V$.
Swapping keys or queries, instead, leads to an improvement over the baseline, reaching a mIoU of $72.3\%$  in the case of keys (that slightly outperform queries). These gains are also justified by reasoning over the operations performed by an attention layer (Sec.~\ref{subsec:cross}).
Finally, the Attention-Mix module shows an improvement of $0.6\%$ in mIoU compared to naive RGBD where the multi-head features are simply summed layer-wise, confirming the necessity of careful feature mixing.
\vspace{-1.5em}
\section{Conclusions}
\label{sec:conclusions}
\vspace{-.5em}
In this work we introduced a strategy that efficiently embeds depth information in the positional encoding of a transformer-based segmentation architecture, improving the final accuracy.
We have also shown how performing cross-attention by acting at the input level can enhance information sharing between parallel network branches, as hinted by the Query-Key-Value interpretation of an attention layer.
These claims were supported by numerical results and ablation studies on the Cityscapes dataset.
In the future, we plan to explore different formulations of positional encodings and different input modalities.

\vspace{-1em}
\bibliographystyle{IEEEtran_nourl}
\bibliography{strings_mid,refs}

\end{document}